\documentclass[twoside]{article}
\usepackage{geometry}
\makeatletter
\renewcommand*{\@fnsymbol}[1]{\ensuremath{\ifcase#1\or \dagger\or \ddagger
\or \mathsection\or \mathparagraph\or \|\or **\or \dagger\dagger \or \ddagger\ddagger
\else\@ctrerr\fi}}
\makeatother

\author{
Ozgur Guldogan\thanks{Currently with Google. This work was performed while at the University of California, Santa Barbara.}
\hspace{0.1cm},
Jackson Kunde\thanks{Currently with Ludo Robotics. This work was performed while at the University of Wisconsin--Madison.}
\hspace{0.1cm},
Kangwook Lee\thanks{Currently with KRAFTON AI and Ludo Robotics. This work was performed while at the University of Wisconsin--Madison.}
\hspace{0.1cm},
Ramtin Pedarsani\thanks{University of California, Santa Barbara.}
}

\title{Multi-Bin Batching for Increasing LLM Inference Throughput}

\usepackage[round]{natbib}
\usepackage{url}

\usepackage{amsmath,amsfonts,bm}

\def\eqref#1{(\ref{#1})}

\def\1{\bm{1}}

\def\rx{{\textnormal{x}}}

\DeclareMathAlphabet{\mathsfit}{\encodingdefault}{\sfdefault}{m}{sl}
\SetMathAlphabet{\mathsfit}{bold}{\encodingdefault}{\sfdefault}{bx}{n}

\newcommand{\E}{\mathbb{E}}

\usepackage{amsmath,amssymb,amsthm,amsfonts,amscd,mathrsfs,mathtools}
\usepackage{graphicx}
\usepackage{textcomp}

\def\BibTeX{{\rm B\kern-.05em{\sc i\kern-.025em b}\kern-.08em
    T\kern-.1667em\lower.7ex\hbox{E}\kern-.125emX}}

\usepackage{graphicx}
\usepackage{amsmath}
\usepackage{amsthm}
\usepackage{amsfonts}
\usepackage[labelformat=simple]{subcaption}

\usepackage{wrapfig}
\usepackage{xcolor}
\usepackage{hyperref}
\usepackage{booktabs}

\usepackage[ruled,vlined]{algorithm2e}

\allowdisplaybreaks

\definecolor{ruddy}{rgb}{1.0, 0.0, 0.16}
\definecolor{gblue}{RGB}{29, 144, 255}
\definecolor{royalblue}{rgb}{0.25, 0.41, 0.88}
\hypersetup{
  colorlinks=true,
  citecolor=royalblue,
  linkcolor=ruddy,
  urlcolor=royalblue}

\newtheorem{definition}{Definition}[section]
\newtheorem{theorem}{Theorem}[section]
\newtheorem*{theorem*}{Theorem}

\newtheorem{corollary}{Corollary}[theorem]
\newtheorem{lemma}[theorem]{Lemma}
\newtheorem{assumption}{Assumption}[section]
\newtheorem{remark}{Remark}[section]

\begin{document}

\date{}

\vspace{-3cm}
\maketitle

\begin{abstract}
As large language models (LLMs) grow in popularity for their diverse capabilities, improving the efficiency of their inference systems has become increasingly critical.
Batching LLM requests is a critical step in scheduling the inference jobs on servers (e.g. GPUs), enabling the system to maximize throughput by allowing multiple requests to be processed in parallel.
However, requests often have varying generation lengths, causing resource underutilization, as hardware must wait for the longest-running request in the batch to complete before moving to the next batch.
We formalize this problem from a queueing-theoretic perspective, and aim to design a control policy which is throughput-optimal under a static-batching framework.
We propose Multi-Bin Batching, a simple yet effective method that can \emph{provably improve LLM inference throughput} under this framework by grouping requests with similar (predicted) execution times into predetermined bins.
Through a combination of theoretical analysis and experiments, including real-world LLM inference scenarios with static and continuous-batching baselines, we demonstrate that multi-bin batching substantially improves throughput over static batching and quantify the remaining gap to native continuous batching under both oracle and estimated length information.
\end{abstract}
\section{Introduction}

Large Language Model (LLM) inference systems are becoming increasingly popular due to their various abilities, such as text generation~\citep{li2024pre}, coding assistance~\citep{chen2021evaluating}, and question answering~\citep{jiang2021can}.
As the demand for LLM inference systems grows, so does the need to optimize their efficiency.

\begin{figure*}[ht]
    \centering
    \begin{subfigure}[t]{0.35\textwidth}
        \centering
        \includegraphics[width=\textwidth]{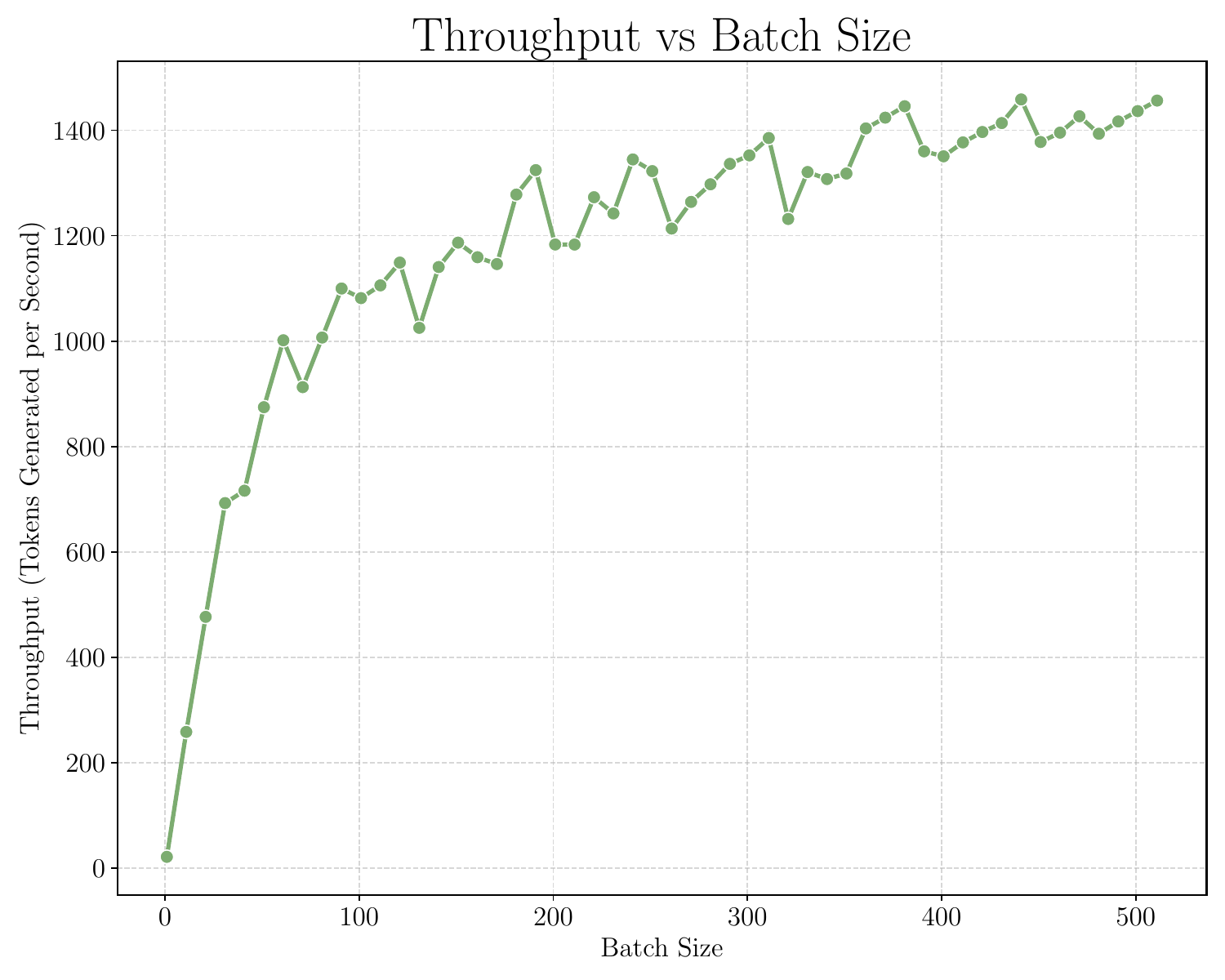}
        \caption{}
        \label{fig:batching_throughput}
    \end{subfigure}
    ~
    \begin{subfigure}[t]{0.60\textwidth}
        \centering
        \includegraphics[width=\textwidth]{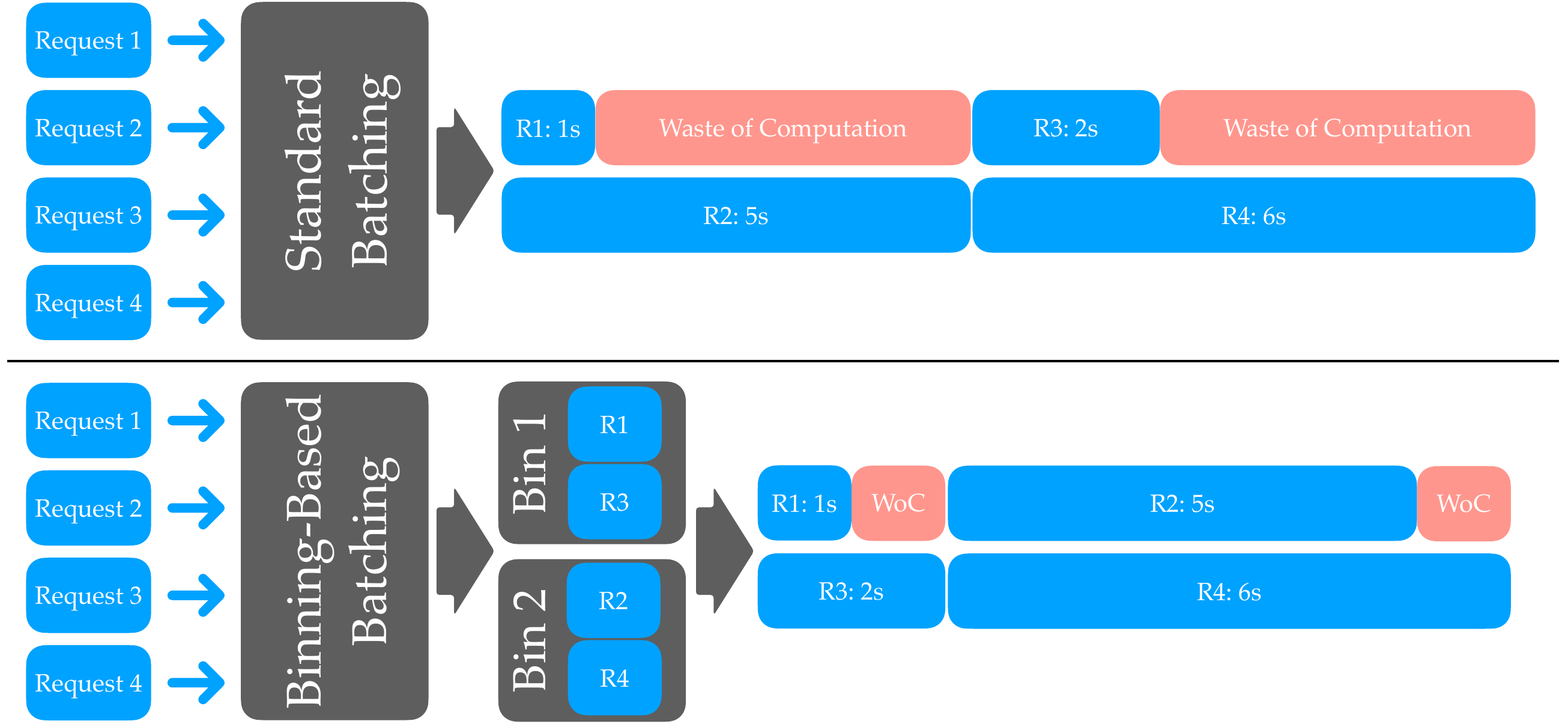}
        \caption{}
        \label{fig:batch_inefficiency}
    \end{subfigure}
\caption{~\subref{fig:batching_throughput} Batch serving improves the throughput for the LLM inference systems. ~\subref{fig:batch_inefficiency} Standard batching causes underutilization of resources due to varying answer sizes.}
\end{figure*}

Several techniques have been proposed to improve the efficiency of LLM inference systems, and \emph{batched inference}~\citep{pmlr-v202-sheng23a, kwon2023efficient, jin2023s} is one of the most promising techniques among them.
With batched inference, multiple requests are processed simultaneously, using the underlying hardware's parallelism to improve throughput.
Figure~\ref{fig:batching_throughput} shows the measured throughput of the Phi-3.5 Mini Instruct model~\citep{abdin2024phi} for various batch sizes on an NVIDIA A100 80G GPU.
Throughput is calculated as the number of total tokens generated across all requests divided by time.

However, batched inference comes with some critical drawbacks.
The execution time of each request depends on the number of tokens generated, which varies across requests.
In standard batched inference systems, a computing unit remains locked until all requests in the batch are completed, leading to resource underutilization when requests within a batch have widely differing execution times.
This inefficiency offsets the throughput gains achieved through parallelism\footnote{With some low-level system engineering, one could potentially allow for dispatch of waiting requests in the queue to a \emph{partially busy} computing node~\citep{yu2022orca}.
We assume such fine-grained request dispatching is not available in this paper.
See Section \ref{sec:relatedwork} for more details.}.

Inspired by this, a natural question arises: can we achieve near-optimal throughput from batched inference?

We propose a novel control policy to optimize batched inference by introducing multi-bin batching that can \emph{provably improve LLM inference throughput in the static-batching model} by grouping requests based on their predicted output lengths.
Instead of placing all requests into a single queue, we create multiple ``bins'', each serving as a waiting area for requests with similar (predicted) output lengths.
This binning policy ensures that requests with comparable execution times are grouped together, minimizing inefficiencies due to varying execution times.
Batches are then formed within each bin and dispatched to a central queue to be processed.

To see why this approach is beneficial, consider the example illustrated in Figure~\ref{fig:batch_inefficiency}, where four requests arrive at nearly the same time, with execution times of 1, 5, 2, and 6 seconds, respectively.
In a standard batching system with a batch size of $B = 2$, the requests would be grouped based on their arrival time, forming two batches: (Request 1: \textit{1s}, Request 2: \textit{5s}) and (Request 3: \textit{2s}, Request 4: \textit{6s}).
The total execution time would be 11 seconds (5 seconds for the first batch and 6 seconds for the second).
On the other hand, two-bin batching (one bin for service time between 1 to 3 seconds and another for those between 4 to 6 seconds) will result in different batches: (Request 1: \textit{1s}, Request 3: \textit{2s}) and (Request 2: \textit{5s}, Request 4: \textit{6s}).
This will reduce the total execution time to 8 seconds, demonstrating how multi-bin batching can improve LLM inference throughput.

Our analysis shows that, within the static-batching framework, increasing the number of bins makes the throughput approach the idealized static-batching capacity by reducing inefficiencies caused by heterogeneous execution times within batches.

On the Dolly dataset workload, estimated-length multi-bin batching improves generated-token throughput by 150.1\% over static batching at $k=8$, but remains 43.0\% below the continuous-batching baseline. The corresponding oracle-length result improves over static batching by 253.1\%, while remaining 19.5\% below continuous batching. These results show that length-aware grouping substantially mitigates static-batching inefficiency, but also that native continuous batching remains a strong practical baseline on this workload.

Although our presentation is motivated by LLM inference, the underlying queueing formulation also applies more broadly to autoregressive batch-serving systems in which heterogeneous completion times create straggler effects within fixed batches.

To summarize, our contributions are as follows:
\begin{itemize}
    \item We propose a novel binning-based scheduling strategy that dynamically groups requests into bins based on output lengths, minimizing inefficiencies from varying execution times and optimizing throughput for LLM inference systems.
    \item We use queueing-theoretical analysis to show that our multi-bin batching strategy achieves asymptotic throughput optimality under the static batching model as the number of bins grows. Additionally, we characterize the number of bins required to achieve a desired throughput improvement.
    \item Our experiments on real LLM inference workloads show that estimated-length multi-bin batching substantially improves generated-token throughput over static batching, reaching 150.1\% improvement on the Dolly dataset workload, while remaining below the continuous-batching baseline.
\end{itemize}

\section{Related Work}
\label{sec:relatedwork}

\paragraph{LLM Inference and Scheduling}
Recent research has focused on optimizing large language model (LLM) inference through scheduling techniques and queueing-theoretic tools.
FastServe~\citep{wu2023fast} utilizes a preemptive scheduling algorithm, skip-join Multi-Level Feedback Queue, to improve job completion time in LLM inference systems.
Prior work on dynamic batching~\citep{Inoue_2021} studies a setting where the system serves up to $B$ jobs depending on queue occupancy and derives closed-form upper bounds for the mean latency.
Slice-level scheduling~\citep{cheng2024slice} splits the maximum output length into slices and serves batches slice by slice, improving memory utilization and response time.
Llumnix~\citep{sun2024llumnix} addresses heterogeneous and unpredictable LLM inference requests through runtime rescheduling across multiple model instances, improving tail latency and resource utilization.
Recent systems further exploit the two-phase structure of LLM inference: Sarathi-Serve~\citep{agrawal2024sarathi} uses chunked prefills and stall-free scheduling to improve the throughput--latency tradeoff, while Splitwise~\citep{patel2023splitwise} separates prompt computation and token generation across different resources to exploit the distinct computational characteristics of prefill and decode.
Queueing-theoretic analysis of low-latency LLM inference~\citep{yang2024queueing} studies an M/G/1 model and shows that maximum-output-token limits and batch-size optimization can significantly reduce latency under token-length variability.
Queue management for SLO-oriented LLM serving~\citep{patke2024queue} introduces grouping strategies based on request length to improve throughput.
Our work is complementary to these studies: rather than modeling general queueing delay in an M/G/1 system or optimizing prefill/decode scheduling, we study a length-aware static-batching control policy, characterize bin boundaries in tractable service-time models, and quantify how finite-bin discretization and prediction errors affect the resulting effective batched service rate.

\paragraph{LLM Serving and Answer Length Estimation}
Several studies have improved LLM inference throughput and latency using response-length or execution-time estimation.
Response-time prediction via prompting~\citep{zheng2024response} asks the model an additional question to estimate response time.
Learning-to-rank based scheduling~\citep{fu2024efficient} predicts the relative ordering of requests by execution time and uses this ranking in a shortest-job-first scheduling algorithm to reduce head-of-line blocking.
Proxy-model-based prediction~\citep{qiu2024efficient} uses a lightweight proxy model to predict execution time and then applies speculative shortest-job-first scheduling to improve serving throughput.
S$^3$~\citep{jin2023s} estimates answer length and uses this estimate to improve memory efficiency, thereby increasing the effective batch size of the system.
Generation-length prediction for LMaaS~\citep{cheng2024enabling} uses input length to predict response length and optimize batch size, improving throughput and reducing response time.
Similarly, SyncIntellects~\citep{lin2024syncintellects} enhances response-length prediction using a transformer-based model and implements QoS-friendly length control, improving both throughput and latency.
These works show that length information can be a useful scheduling signal in LLM serving. Our method uses this signal differently: instead of selecting a job order or batch size, it partitions requests into length-aware bins before batch formation.

\paragraph{LLM Inference Optimization} Recent studies have focused on optimizing the inference efficiency of LLM models through various techniques.
Quantization has emerged as a key approach to reduce the memory footprint and improve the inference efficiency of LLM models.
Methods like LLM.Int8()~\citep{dettmers2022gpt3}, GPTQ~\citep{frantar2023optq}, SmoothQuant~\citep{xiao2023smoothquant}, and AWQ~\citep{lin2024awq} have demonstrated effective weight quantization techniques for LLM models, while QLoRA~\citep{dettmers2024qlora} combines quantization with parameter-efficient fine-tuning.
Memory management innovations such as PagedAttention~\citep{kwon2023efficient} have significantly improved the serving throughput.
KV cache optimizations, including compression techniques like Gear~\citep{kang2024gear}, have further improved the memory efficiency of LLM inference systems.
Systems such as FastServe~\citep{wu2023fast} and FlexGen~\citep{pmlr-v202-sheng23a} have integrated these techniques to create comprehensive LLM serving solutions.

\paragraph{Continuous Batching}
Continuous batching, also known as iteration-level scheduling, allows completed requests to leave a batch and new requests to enter during decoding, rather than requiring a fixed batch to remain locked until all requests finish~\citep{yu2022orca}.
Modern serving systems such as Orca~\citep{yu2022orca}, vLLM/PagedAttention~\citep{kwon2023efficient}, and SGLang~\citep{zheng2023sglang} use this style of scheduling or related mechanisms to improve utilization and reduce the head-of-line blocking present in static batching.
At the same time, continuous-batching systems must manage heterogeneous prefill and decode phases, KV-cache constraints, and scheduling hyperparameters that affect the balance between throughput and latency.
This distinction is especially important because prefill and decode have different computational and memory characteristics, and recent systems such as Sarathi-Serve~\citep{agrawal2024sarathi} and Splitwise~\citep{patel2023splitwise} explicitly exploit this two-phase structure.
Our theoretical analysis does not attempt to characterize iteration-level continuous batching. Instead, it provides a tractable static-batching framework that isolates the effect of output-length heterogeneity. In the experiments, we therefore include the backend's native continuous-batching behavior as a strong practical reference baseline. The resulting interpretation is that multi-bin batching is a length-aware grouping principle that may complement modern serving systems, for example through queue classes, or admission-control classes rather than a replacement for continuous batching.

\section{Problem Setup and the Multi-Bin Batching Algorithm}
We begin by introducing the system model and key assumptions that form the foundation of our analysis.
This model represents a typical LLM inference system as a queueing system with specific characteristics.
Following this, we will propose our novel batching algorithm, which leverages a multi-binning approach to optimize request processing.
The model studied in this section is a stylized static-batching framework: it is designed to isolate the effect of service-time, and in particular decode/output-length, heterogeneity on batch efficiency rather than to capture all costs in a production LLM serving system.
\begin{assumption}\label{asm:queueing_system}
    The LLM inference system is a single-server queueing system with an infinite queue length capacity.
    The system receives requests from a Poisson process with rate $\lambda$.
\end{assumption}
Assumption~\ref{asm:queueing_system} is a standard assumption in queueing theory, and it is well-suited for LLM inference systems since requests are typically generated by users in a random manner.
Although this analysis assumes a single-server system, extension to multi-server systems is rather straightforward.

The system forms batches of size $B$, and the serving time of a batch is determined by the maximum of the serving times of its requests.
For the purposes of this paper, we assume a ``first formed batch, first served" policy, where batches are processed in the order they are fully formed, regardless of when its constituent requests arrived.
Alternative policies, such as ``shortest batch, first served", could also be considered depending on specific system requirements.

\begin{assumption}\label{asm:service_time_uniform}
    The service time of each request is independent and identically distributed (i.i.d.) with a uniform distribution in the range $[l_{\min}, l_{\max}]$, i.e., $l \sim \mathcal{U}(l_{\min}, l_{\max})$.
\end{assumption}
We make this assumption to simplify the analysis, and it is justified since LLM answer lengths typically fall within a specific range due to maximum token length limitations.
We also extend our analysis to the case where the service time is exponentially distributed, which can be found in the Section~\ref{sec:exp_service_time}.
The uniform and exponential assumptions should be interpreted as analytically tractable service-time models, not as claims that real LLM inference times exactly follow these distributions. In autoregressive generation, decode time is often closely related to the number of generated tokens, so the service-time variable can be viewed as an abstraction of output-length-driven decode cost. The analysis is therefore most directly applicable to decode-dominant workloads, or to systems in which prefill and decode are separated. It does not explicitly model prefill time, input-length heterogeneity, or the interaction between prefill and decode costs; these factors are included only in the end-to-end empirical measurements in Section~\ref{sec:llm-experiments}.

In our theoretical analysis, we assume that the system always forms batches of size $B$ and then starts processing them.
However, in real systems, there could be a parameter that specifies the maximum time a batch can wait before it is processed.
This way the system can ensure that the latency of a request does not exceed a certain threshold and the quality of service is maintained.

\subsection{Multi-Bin Batching Algorithm}

We propose a novel scheduling policy that aims to improve the throughput of LLM inference systems. The key idea is to group requests into $k$ bins based on their service times before forming batches within each bin.
This binning procedure reduces inefficiencies by ensuring that requests with similar execution times are batched together.

\begin{algorithm}[h]
\LinesNumbered
\caption{Multi-Bin Batching with $k$-bins}
\KwIn{Bin boundaries, $[l_{i-1}, l_i]$, $i = 1, \ldots, k$-bins, batch size $B$, and serving policy}
\For{each incoming request}{
    Estimate its service time $l$\;
    Assign the request to bin $i$ where $l_{i-1} \leq l < l_i$\;
}
\For{each bin}{
    Form batches of size $B$ when available\;
    Add completed batches to the service queue\;
}
Serve batches from the service queue based on the serving policy provided\;
The serving time of a batch is the maximum of the serving times of the requests in the batch\;
\end{algorithm}

\begin{wrapfigure}{r}{0.42\textwidth}
    \centering
    \includegraphics[width=0.40\textwidth]{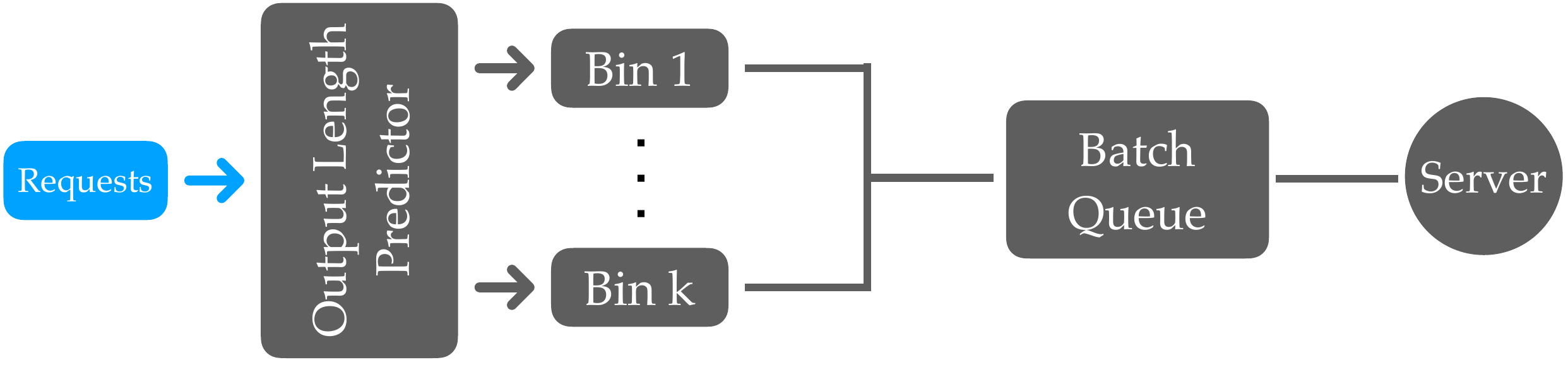}
\caption{k-Bin Batching}
    \label{fig:binning_system}
\end{wrapfigure}

The multi-bin batching algorithm is illustrated in Figure~\ref{fig:binning_system}.
This algorithm works by first dividing the range of possible service times into $k$ bins with boundaries determined by prior knowledge or system profiling.
Incoming requests are assigned to bins based on their estimated service times and grouped into batches of size $B$.
Completed batches are added to a service queue and processed using the ``first completed batch, first served" policy, the default for this paper.
By batching requests with similar service times together, this method reduces the serving time of each batch, and ensures efficient processing across all bins.

\section{Throughput Analysis}

In this section, we present a queueing-theoretical analysis to show that our multi-bin batching policy improves the throughput of LLM inference systems within the static-batching framework introduced above.
To analyze the throughput of the system, we first derive the optimal decision boundaries for each bin in our batching system.
Using these boundaries, we calculate the expected service time of a batch of requests processed under the multi-bin batching policy.
Throughout this section, ``throughput'' refers to the saturated request processing rate of the static-batching, i.e., the number of requests completed per unit service time. This is a service-rate quantity for the batched server and should be distinguished from the maximum stable arrival rate of a full queueing system with finite waiting effects.

\begin{definition}[Throughput under static batching]\label{def:effective_batched_service_rate}
    For a saturated static-batching server, the throughput is the ratio of the batch size to the expected service time. Specifically, it can be written as
    \begin{equation}
        \text{Throughput} = \frac{B}{\E[t_{\text{service}}]},
    \end{equation}
    where $B$ is the batch size and $\E[t_{\text{service}}]$ is the expected service time of a batch of $B$ requests.
\end{definition}

\begin{remark}
    Within this static-batching framework, the idealized upper bound on throughput is achieved when the service time of a batch, $t_{\text{service}}$, equals the average service time of a single request rather than the maximum service time within the batch.
    In this case, the expected throughput becomes, $\text{Throughput}_{\text{max}} = \frac{B}{\E[t]},$ where $\E[t]$ is the mean service time of a single request.
    This scenario represents the idealized case of perfect parallelism without any delays due to batching overhead.
    This bound is specific to the fixed-batch model analyzed here and is not intended as a universal upper bound for continuous-batching systems, which can pipeline request admission and completion at a finer granularity.
\end{remark}

For our multi-bin batching system, we can derive the expected service time of a batch of $B$ requests as follows:
\begin{equation}
    \E[t_{\text{service, k}}] = \sum_{i=1}^k \Pr(\text{bin}=i)\E\left[\max_{j\in[B]}\rx_{j}|\text{bin}=i\right],
\end{equation}
where $\Pr(\text{bin}=i)$ is the probability that a batch served by the system is in bin $i$, and the expected service time of a batch of $B$ requests from bin $i$ is denoted by $\E\left[\max_{j\in[B]}\rx_{j}|\text{bin}=i\right]$.
We also denote it as $\E[t_{\text{service, k}}]$ to emphasize that it is the expected service time of multi-bin batching system with $k$ bins.
Then, the first step is to determine the optimal decision boundaries for each bin in the multi-bin batching system for a fixed number of bins $k$, as provided by the following lemma.
\begin{lemma}\label{lem:optimal_decision_boundary_uniform}
    Under Assumption~\ref{asm:service_time_uniform} and a fixed number of bins $k$, the throughput is maximized when each bin has equal probability mass, and the decision boundaries are determined as follows:

    \begin{equation}\label{eq:optimal_decision_boundary_uniform}
        l_i = l_{\min} + \frac{i}{k}(l_{\max} - l_{\min}), \quad i\in[k].
    \end{equation}
    and where $l_0 = l_{\min}$.
\end{lemma}
\begin{proof}
    We begin by defining the expected service time of a request in bin $i$ as follows,
    \begin{equation}
        \E\left[\max_{j\in[B]}\rx_{j}|\text{bin}=i\right]
        = \frac{1}{B+1}l_{i-1} + \frac{B}{B+1}l_{i}.
    \end{equation}
    Since, the service time of a request in bin $i$ is uniformly distributed in the range $[l_{i-1}, l_{i}]$, the expected value of the maximum of $B$ uniform random variables in the range $[l_{i-1}, l_{i}]$ is well-known and can be computed easily.
    Then, the expected service time of the system is given by,
    \begin{equation}
        \E\left[t_{\text{service, k}}\right] = \sum_{i=1}^k \Pr(\text{bin}=i)\E\left[\max_{j\in[B]}\rx_{j}|\text{bin}=i\right]
    \end{equation}
    It can be written as,
    \begin{equation}
        \E\left[t_{\text{service, k}}\right] = \sum_{i=1}^k \frac{l_{i} - l_{i-1}}{l_{\max} - l_{\min}}\left(\frac{1}{B+1}l_{i-1} + \frac{B}{B+1}l_{i}\right).
    \end{equation}
    For $k$ bins, we have $k-1$ decision boundaries, and $l_{0}=l_{\min}$ and $l_{k}=l_{\max}$.
    We can denote the expected service time of the system as a function of $\boldsymbol{l}=[l_{1}, l_{2}, \ldots, l_{k-1}]$ as follows,
    \begin{equation}
        f_{k}(\boldsymbol{l}) = \sum_{i=1}^k \frac{l_{i} - l_{i-1}}{l_{\max} - l_{\min}}\left(\frac{1}{B+1}l_{i-1} + \frac{B}{B+1}l_{i}\right).
    \end{equation}
    We can compute the partial derivative of $f_{k}(\boldsymbol{l})$ with respect to $l_{i}, i\in[k-1]$ as follows,
    \begin{equation}\label{eq:partial_derivative_f_k}
        \frac{\partial f_{k}(\boldsymbol{l})}{\partial l_{i}} = -\frac{B-1}{B+1}\frac{l_{i+1}}{l_{\max} - l_{\min}} \\+ \frac{2(B-1)}{B+1}\frac{l_{i}}{l_{\max} - l_{\min}} - \frac{B-1}{B+1}\frac{l_{i-1}}{l_{\max} - l_{\min}}.
    \end{equation}
    Then, the second-order partial derivative of $f_{k}(\boldsymbol{l})$ with respect to $l_{i}$ and $l_{j}, i,j\in[k-1]$ is given by,
    \begin{equation}
        \frac{\partial^{2} f_{k}(\boldsymbol{l})}{\partial l_{i}\partial l_{j}} = \begin{cases}
            \frac{2(B-1)}{B+1}\frac{1}{l_{\max} - l_{\min}} & \text{if } i=j,\\
            -\frac{B-1}{B+1}\frac{1}{l_{\max} - l_{\min}} & \text{if } |i-j|=1,\\
            0 & \text{otherwise}.
        \end{cases}
    \end{equation}
    The Hessian matrix of $f_{k}(\boldsymbol{l})$ is a tridiagonal matrix in the form of,
    \begin{equation}
        \nabla^{2} f_{k}(\boldsymbol{l}) =c\begin{bmatrix}
            2 & -1 & 0 & \cdots & 0 & 0\\
            -1 & 2 & -1 & \cdots & 0 & 0\\
            0 & -1 & 2 & \cdots & 0 & 0\\
            \vdots & \vdots & \vdots & \ddots & \vdots & \vdots\\
            0 & 0 & 0 & \cdots & 2 & -1\\
            0 & 0 & 0 & \cdots & -1 & 2
        \end{bmatrix},
    \end{equation}
    where
    \begin{equation}
        c = \frac{B-1}{B+1}\frac{1}{l_{\max} - l_{\min}}.
    \end{equation}
    The determinant of the Hessian matrix can be computed via the recursive formula for the determinant of a tridiagonal matrix as follows,
    \begin{equation}
        \det(\nabla^{2} f_{k}(\boldsymbol{l})) = k\left(\frac{(B-1)}{(B+1)(l_{\max}-l_{\min})}\right)^{k-1} > 0
    \end{equation}
    Since $k>1$ and $B>1$, and since the tridiagonal matrix with diagonal entries $2$ and off-diagonal entries $-1$ is positive definite, the Hessian matrix is positive definite.
    Therefore, the function $f_{k}(\boldsymbol{l})$ is convex with respect to $l_{1}, l_{2}, \ldots, l_{k-1}$.
    Then, one can solve Equation~\eqref{eq:partial_derivative_f_k} for $l_{i}$ by setting the partial derivative to zero, i.e., $\frac{\partial f_{k}(\boldsymbol{l})}{\partial l_{i}}=0$.
    It can be seen that the optimal decision boundaries are given by,
    \begin{equation}
        l_{i} = l_{\min} + \frac{i}{k}(l_{\max} - l_{\min}) \quad \forall i\in[k-1].
    \end{equation}
    This completes the proof.
\end{proof}

Given the optimal decision boundaries in Lemma~\ref{lem:optimal_decision_boundary_uniform}, for a fixed number of bins $k$, we can have the following theorem for the expected throughput of the system.
\begin{theorem}\label{thm:throughput_k_bin}
    Under Assumption~\ref{asm:service_time_uniform}, the expected throughput of the $k$-bin batching is

    \begin{equation}\label{eq:throughput_k_bin}
        \text{Throughput}_{k} = \frac{B}{\E[t_{\text{service, k}}]} = \frac{B}{\frac{l_{\max} + l_{\min}}{2} + \frac{1}{k}l_B},
    \end{equation}
    where
    \begin{equation*}l_B=
        \frac{B}{B+1}l_{\max} + \frac{1}{B+1}l_{\min} - \frac{l_{\max} + l_{\min}}{2}
    \end{equation*}

    and it is an increasing function of the number of bins $k$.
\end{theorem}

\begin{proof}
    The expected service time of a batch of $B$ requests is,
    \begin{align}\label{eq:service_time_k_bin}
        \E[t_{\text{service, k}}] &= \sum_{i=1}^k \Pr(\text{bin}=i)\E\left[\max_{j\in[B]}\rx_{j}|\text{bin}=i\right] \\&= \sum_{i=1}^k \frac{1}{k}\left(\frac{B}{B+1}l_i + \frac{1}{B+1}l_{i-1}\right),
    \end{align}
    because each bin has equal probability mass, and the service time of a batch of requests follows from the uniform distribution in the range $[l_{i-1}, l_i]$.
    If we substitute the optimal decision boundaries in Equation~(\ref{eq:optimal_decision_boundary_uniform}) into Equation~(\ref{eq:service_time_k_bin}), we can derive the expected service time of a batch of $B$ requests with multi-bin batching as follows,
    \begin{align}\label{eq:throughput_k_bin_deriv}
        \E[t_{\text{service, k}}] &= \frac{1}{k}\sum_{i=1}^k\frac{B}{B+1}\left(l_{\min} + \frac{i}{k}(l_{\max} - l_{\min})\right) + \frac{1}{B+1}\left(l_{\min} + \frac{i-1}{k}(l_{\max} - l_{\min})\right)\nonumber \\
        &= \frac{1}{k}\frac{B}{B+1}\frac{k+1}{2}(l_{\max} - l_{\min}) + \frac{1}{k}\frac{1}{B+1}\frac{k-1}{2}(l_{\max} - l_{\min}) + l_{\min}\nonumber \\
        &= \frac{l_{\max} + l_{\min}}{2} + \frac{1}{k}\left(
        \frac{B}{B+1}l_{\max} + \frac{1}{B+1}l_{\min} - \frac{l_{\max} + l_{\min}}{2}\nonumber
        \right)
    \end{align}

    Then, the expected throughput of the system with multi-bin batching with $k$ bins is,
    \begin{equation}
        \text{Throughput}_{k} = \frac{B}{\E[t_{\text{service, k}}]} = \frac{B}{\frac{l_{\max} + l_{\min}}{2} + \frac{1}{k}l_B},
    \end{equation}
    and it can be observed that it is an increasing function of the number of bins $k$ since the denominator is decreasing with respect to $k$.
\end{proof}

\begin{remark}
    The standard batching system is a special case of the multi-bin batching system with $k=1$.
    If we substitute $k=1$ into Equation~(\ref{eq:throughput_k_bin}), we can derive the expected throughput of the system with standard batching.
    Since the expected throughput of the system with the multi-bin batching is an increasing function of the number of bins $k$, the throughput of the system with our multi-bin batching system is higher than the standard batching system.
    Hence, the multi-bin batching system can improve the throughput of the system.
\end{remark}
\begin{remark}
The expected throughput of the system with multi-bin batching increases with the number of bins $k$ and converges to the maximum capacity of the system as $k\to\infty$:
\begin{equation}
        c_{\max} = \lim_{k\to\infty}\text{Throughput}_{k} = \frac{B}{\frac{l_{\max} + l_{\min}}{2}}.
\end{equation}
With infinitely many bins, the binning becomes so fine-grained that the batch service time equals that of a single request, achieving the optimal throughput of static batching framework asymptotically.
\end{remark}

In the following theorem, we derive the minimum number of bins $k$ required to achieve a target throughput below the maximum capacity.
\begin{theorem}\label{thm:min_k}
    Under Assumptions~\ref{asm:queueing_system}, and~\ref{asm:service_time_uniform}, for any $\epsilon>0$, the desired throughput of the system $c_{\max}-\epsilon$ can be achieved by the multi-bin batching system with $k$ bins, where $k$ is the smallest integer satisfying the condition
    \begin{align}\label{eq:min_k}
        k &\geq \left\lceil\frac{(c_{\max}-\epsilon)l_B}{\epsilon\frac{l_{\max} + l_{\min}}{2}}\right\rceil = O\left(\frac{1}{\epsilon}\right).
    \end{align}
\end{theorem}
\begin{proof}
    The desired throughput of the system is $c_{\max}-\epsilon$.
    From Theorem~\ref{thm:throughput_k_bin}, the expected throughput of the system with multi-bin batching with $k$ bins is,
    \begin{equation}
        \text{Throughput}_{k} = \frac{B}{\frac{l_{\max} + l_{\min}}{2} + \frac{1}{k}l_B}.
    \end{equation}
    where $l_B = \frac{B}{B+1}l_{\max} + \frac{1}{B+1}l_{\min} - \frac{l_{\max} + l_{\min}}{2}$.
    Then, we can find the smallest integer $k$ that satisfies the following condition,
    \begin{equation}
        c_{\max}-\epsilon \leq  \frac{B}{\frac{l_{\max} + l_{\min}}{2} + \frac{1}{k}l_B}.
    \end{equation}
    We can solve the above inequality for $k$ to find the smallest integer $k$ that satisfies the desired throughput of the system,
    \begin{equation}
        (c_{\max}-\epsilon)\left[\left(\frac{l_{\max} + l_{\min}}{2}\right) + \frac{1}{k}l_B\right] \leq B.
    \end{equation}
    It can be simplified as follows,
    \begin{equation}
        B-\epsilon\left(\frac{l_{\max} + l_{\min}}{2}\right) + (c_{\max}-\epsilon)\frac{1}{k}l_B \leq B.
    \end{equation}
    This implies,
    \begin{equation}
        k \geq \frac{(c_{\max}-\epsilon)l_B}{\epsilon\frac{l_{\max} + l_{\min}}{2}}.
    \end{equation}
    Therefore, the smallest integer $k$ that satisfies the desired throughput of the system is given as in the statement of the theorem.
\end{proof}

\begin{figure}[ht]
    \centering
    \includegraphics[width=0.45\textwidth]{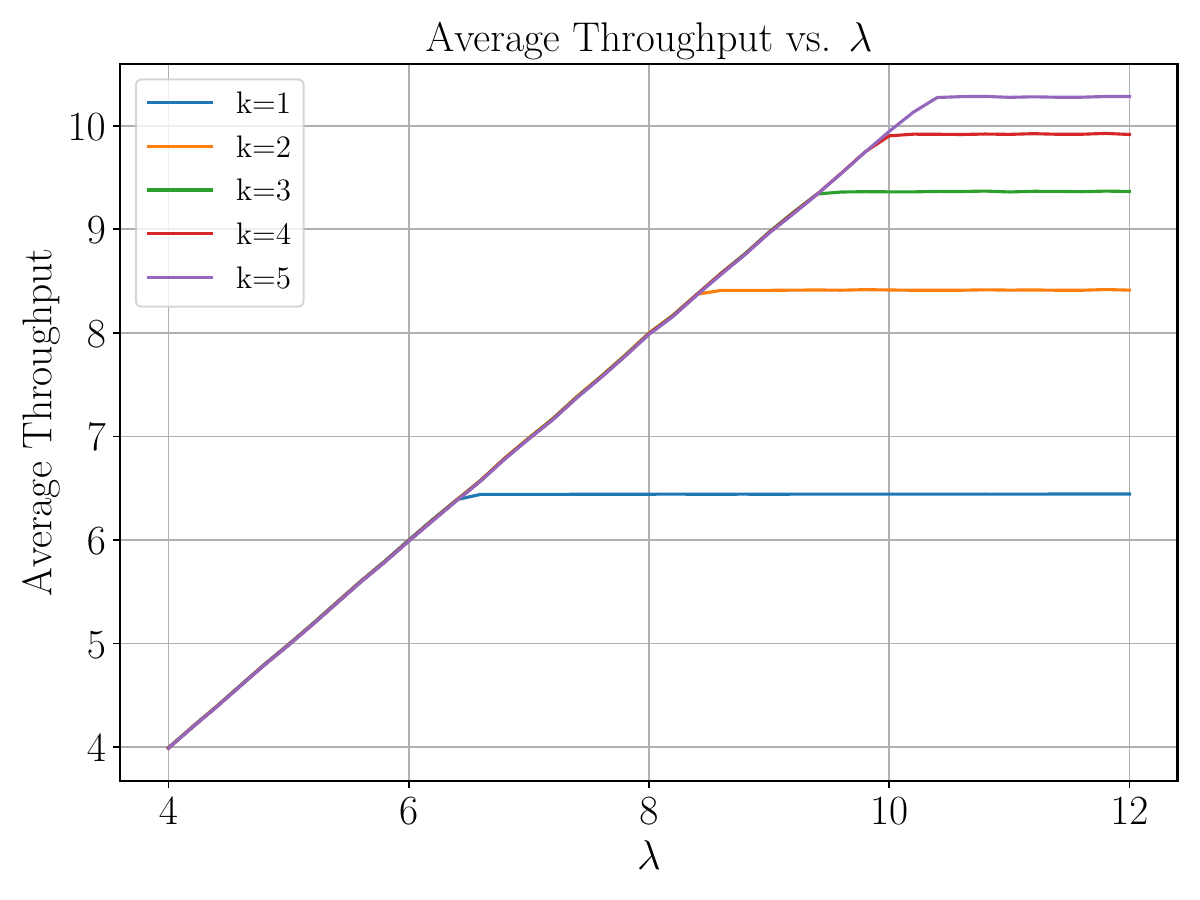}
\caption{Average throughput of the system with varying numbers of bins $k$ as a function of the arrival rate $\lambda$.}
    \label{fig:average_throughput_vs_lambda}
\end{figure}

Figure~\ref{fig:average_throughput_vs_lambda} shows the average throughput of the system with varying numbers of bins $k$ as a function of the arrival rate $\lambda$.
In this figure, we assumed that the batch size $B=128$, the minimum service time $l_{\min}=1$, and the maximum service time $l_{\max}=20$, resulting a maximum theoretical capacity of $c_{\max} = \frac{128}{\frac{20+1}{2}} \approx 12.3$.
We submit 128000 requests to the system and measure the time taken to process all the requests.
We run the simulations for 10 times and report the average throughput of the system.
Then, we calculate the average throughput of the system as the number of requests processed per unit time.
The results show that throughput increases with the number of bins $k$, approaching the maximum capacity when $k=5$.
However, increasing the number of bins introduces a trade-off: as $k$ grows, the time required to construct batches also increases, potentially leading to higher latency.
In the next section, we evaluate the latency of the system with multi-bin batching and compare it with the standard batching system.

\section{Latency Analysis}

The latency of a request is defined as the total time from when it enters the system until it is completely served.
The latency of a request consists of two components: the queuing time and the service time.
In the previous section, we analyzed the expected service time of a request for our multi-bin batching scheduling policy.
In this section, we focus on the queuing time, which can be further decomposed into two components: the time spent waiting to form the current batch and the time spent waiting for the current batch to start processing.
In this analysis, we focus on the time spent waiting to form the current batch, which provides a lower bound for the latency of a request.
Thus, the analysis in this section should be interpreted as an idealized batch-formation latency calculation, rather than a complete end-to-end latency model for a loaded single-server system.

To derive this theoretical lower bound for latency, we assume ideal conditions where batches are processed immediately upon formation.
\begin{assumption}\label{asm:infinite_servers}
    The number of servers in the system is infinite. Therefore, any batch that is ready for processing begins immediately without delay.
\end{assumption}

This assumption allows us to approximate latency in an idealized setting, where the waiting time for batches to start processing is negligible.
Equivalently, Assumption~\ref{asm:infinite_servers} intentionally removes server-side waiting after a batch has formed, so the resulting expression isolates only the batch-formation component and the service-time component.
While this lower bound holds across all load conditions, it closely approximates the latency in underloaded systems.
We denote the expected latency of a request as $\mathbb{E}[t_{\text{latency}}]$.
The following lemma provides the expected latency of a request in our system with the assumption of infinite servers.
\begin{lemma}\label{lemma:latency}
    Under Assumptions~\ref{asm:service_time_uniform}, and~\ref{asm:infinite_servers}, and given the arrival rate $\lambda$ and $k$-bins with equal probability mass, the expected latency of a request is given by
    \begin{equation*}
        \mathbb{E}[t_{\text{latency}}] = \frac{l_{\max} + l_{\min}}{2} + \frac{B-1}{2\lambda}k + \frac{1}{k}l_B.
    \end{equation*}
\end{lemma}

The proof of Lemma~\ref{lemma:latency} is given in Appendix~\ref{sec:proof_latency}.
The proof utilizes the fact that the arrival process is Poisson and the effective arrival rate of each bin is $\lambda/k$.
The inverse dependence on $\lambda$ in Lemma~\ref{lemma:latency} reflects the fact that higher arrival rates reduce the time needed to collect enough requests within each bin. It should not be interpreted as a claim that end-to-end latency necessarily decreases with traffic in a practical single-server system, where server-side queueing delay can dominate.

\begin{figure}[ht]
    \centering
    \includegraphics[width=0.45\textwidth]{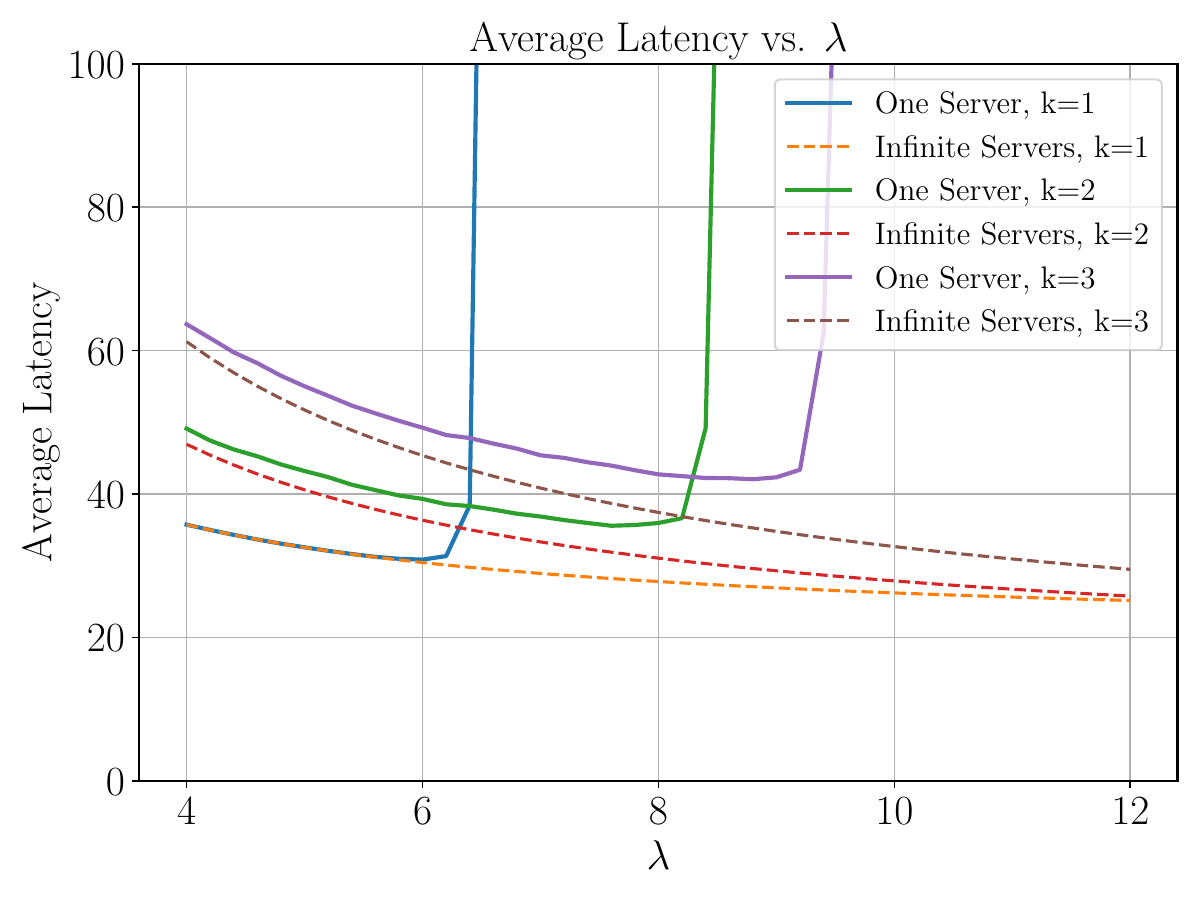}
\caption{The latency of a request vs the arrival rate $\lambda$ for different $k$ values.}
    \label{fig:average_latency_vs_lambda}
\end{figure}

In Figure~\ref{fig:average_latency_vs_lambda}, we plot the average latency of a request as a function of the arrival rate $\lambda$ for different $k$ values.
We use the parameters $l_{\min}=1$, $l_{\max}=20$, $B=128$, and $k=1, 2, 3$.
We submit 128000 requests to the system and measure the latency of each request.
We run the simulations for 10 times and report the average.
It can be seen that our Lemma~\ref{lemma:latency} provides a close approximation of the latency of a request in our system when the arrival rate is low.
In the idealized batch-formation component captured by Lemma~\ref{lemma:latency}, increasing the arrival rate reduces the time needed to form a batch. In a one-server system, however, this effect can be offset by server-side queueing as the system approaches its service-rate limit.
Overall, these results illustrate that multi-bin batching can improve throughput relative to standard static batching, but the number of bins should be selected with the associated batch-formation latency in mind.
More generally, the number of bins $k$ controls a tradeoff between latency and throughput: increasing $k$ reduces service-time waste within a batch, but also decreases the arrival rate into each bin and can increase the time required to form a full batch. Hence, $k$ should be viewed as a workload and latency dependent tuning parameter. Larger values of $k$ are most attractive in high-load or throughput-oriented regimes, while smaller values may be preferable in low-load or latency-sensitive regimes. A full end-to-end latency--throughput Pareto analysis is an important direction for future work.

\section{Analysis under Exponentially Distributed Service Time}\label{sec:exp_service_time}

In this section, we provide the expected service time of a batch of $B$ requests when the service time of each request is exponentially distributed with rate $\mu$.
As in the uniform case, the exponential model is used as an analytically tractable service-time distribution for studying the effect of length heterogeneity under static batching; it is not intended as an empirical claim about the exact distribution of LLM inference times.
Hence, here we make the following assumption:
\begin{assumption}\label{assumption:exp_service_time}
    The service time of each request is independent and identically distributed (i.i.d.) with an exponential distribution with rate $\mu$, i.e., $l \sim \text{Exp}(\mu)$.
\end{assumption}

Then, for our multi-bin batching system, we need to decide the optimal decision boundaries to minimize the expected service time of a batch of $B$ requests.
One can utilize the order statistics of the truncated exponential distribution~\citep{joshi1978recurrence} to derive the expected service time of a batch of $B$ requests with $k$ bins.
However, the exact values of truncated exponential order statistics are not easy to compute.
Therefore, we use a simpler approach to derive an upper bound on the expected service time of a batch of $B$ requests with $k$ bins.
For the bins before the last bin, we can upper bound the expected service time of a batch of $B$ requests as the decision boundary of that bin, i.e., for bin $i$, the expected service time of a batch of $B$ requests is upper bounded by $l_{i}$.
For the last bin, the exact expected service time of a batch of $B$ requests is known and it is $l_{k-1}+\frac{H_B}{\mu}$, where $H_B$ is the $B$-th harmonic number.
Then, we have the following lemma.
\begin{lemma}\label{lem:exp_service_time}
    Under Assumption~\ref{assumption:exp_service_time}, the expected service time of a batch of $B$ requests with $k$ bins is upper bounded by,
    \begin{equation}
    \E[t_{\text{service, k}}] \leq \sum_{i=1}^{k-1} \Pr(\text{bin}=i)l_{i} + \Pr(\text{bin}=k)\left(l_{k-1} + \frac{H_B}{\mu}\right).
    \end{equation}
    and this upper bound is minimized when the decision boundaries are set as,
    \begin{equation}
    l_{i} = \frac{1}{\mu}\sum_{j=1}^{i}\log(L_{k-j}) \quad \forall i\in[k-1]
    \end{equation}
    where $L_m$ is defined recursively as:
    \begin{equation}
    L_m = \begin{cases}
    H_B & \text{if } m = 1 \\
    1+\log(L_{m-1}) & \text{if } m > 1
    \end{cases}
    \end{equation}
\end{lemma}

The proof of Lemma~\ref{lem:exp_service_time} is given in Appendix~\ref{sec:proof_exp_service_time}.
The proof bounds the expected service time by breaking it into contributions from each bin and reformulating the problem using a logarithmic transformation.
It then shows the reformulated upper bound is convex via its Hessian so that setting its gradient to zero yields the optimal decision boundaries.

Given the optimal decision boundaries, the service-time upper bound in Lemma~\ref{lem:exp_service_time} yields the following lower bound on throughput.
\begin{corollary}\label{cor:exp_service_time}
    Under Assumption~\ref{assumption:exp_service_time} and the optimal decision boundaries in Lemma~\ref{lem:exp_service_time}, the expected service time satisfies
    \begin{equation}
        \E[t_{\text{service, k}}] \leq \frac{1}{\mu} f(\boldsymbol{q}),
    \end{equation}
    and therefore the throughput satisfies
    \begin{equation}
        \text{Throughput}_{k}
        =
        \frac{B}{\E[t_{\text{service, k}}]}
        \geq
        \frac{B\mu}{f(\boldsymbol{q})},
    \end{equation}
    where
    \begin{equation}
        f(\boldsymbol{q}) =
        \sum_{i=1}^{k-1}
        \frac{L_{k-i}-1}{\prod_{j=1}^{i} L_{k-j}}
        \cdot
        \sum_{j=1}^{i} \log(L_{k-j})
        +
        \frac{1}{\prod_{j=1}^{k-1} L_{k-j}}
        \left(
        \sum_{j=1}^{k-1} \log(L_{k-j}) + H_B
        \right),
    \end{equation}
    and $L_m$ is defined recursively as in Lemma~\ref{lem:exp_service_time}.
\end{corollary}

\begin{proof}
    The proof of corollary follows from the optimal decision boundaries in Lemma~\ref{lem:exp_service_time} and the expected service time of a batch of $B$ requests with $k$ bins.
\end{proof}

\begin{figure}[h]
    \centering
    \begin{subfigure}[t]{0.48\textwidth}
        \centering
        \includegraphics[width=\textwidth]{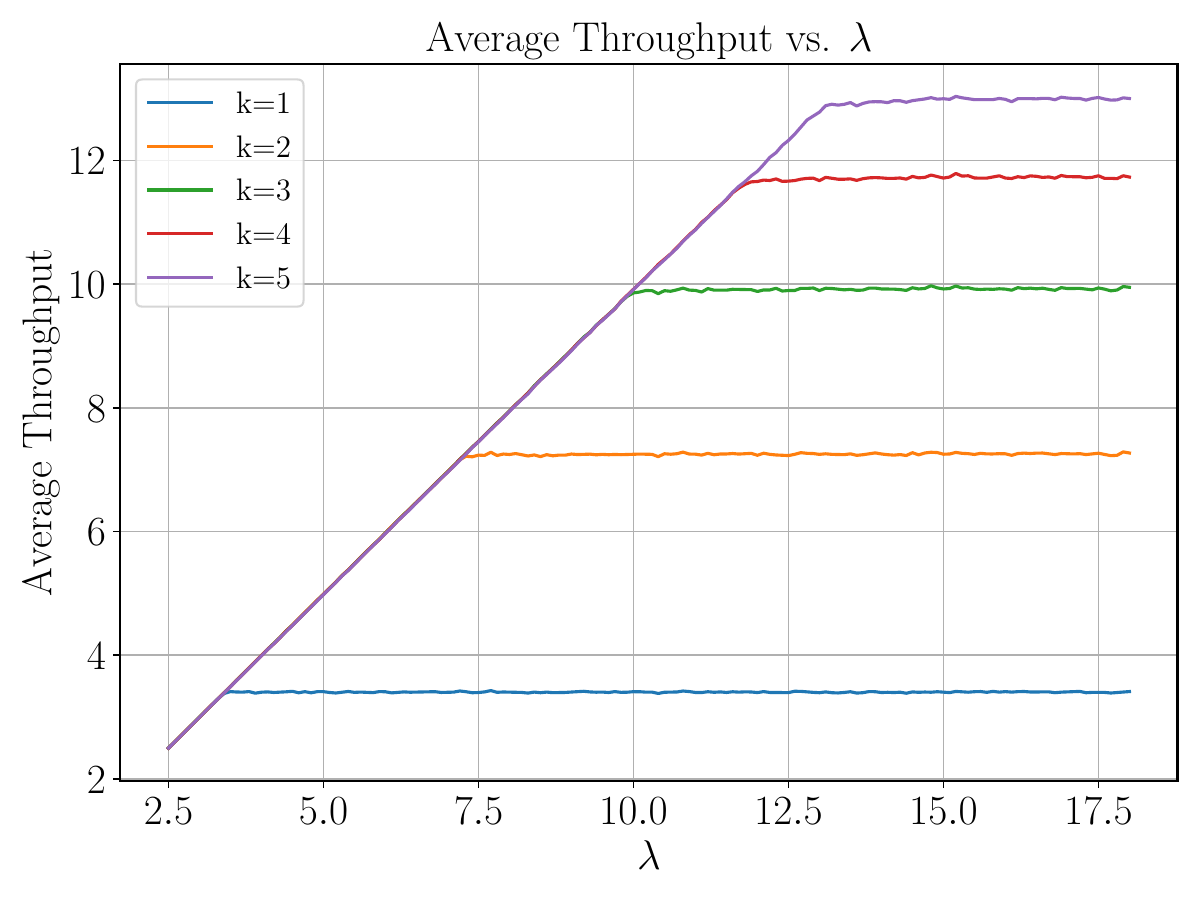}
\caption{Throughput}
        \label{fig:exp_service_time_throughput}
    \end{subfigure}
    \hfill
    \begin{subfigure}[t]{0.48\textwidth}
        \centering
        \includegraphics[width=\textwidth]{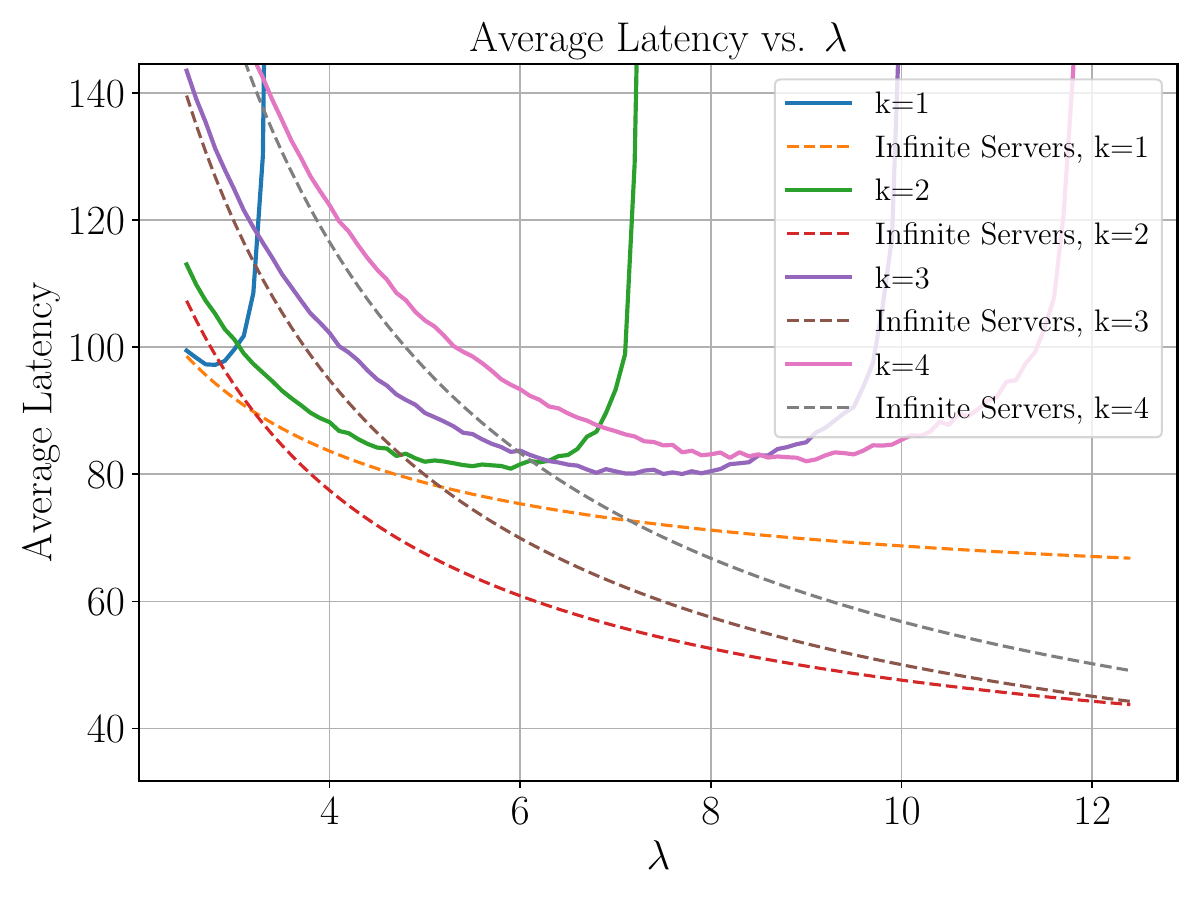}
\caption{Latency}
        \label{fig:exp_service_time_latency}
    \end{subfigure}
    \caption{Throughput and expected latency with respect to the arrival rate $\lambda$ for exponentially distributed service times.}
    \label{fig:exp_service_time_results}
\end{figure}

Similarly to the uniform distribution case, we provide numerical results for the exponentially distributed service time case.
In Figure~\ref{fig:exp_service_time_throughput} and~\ref{fig:exp_service_time_latency}, we provide the throughput and expected latency of the system with respect to the arrival rate $\lambda$ for different values of $k$.
We assume that the service time of each request is exponentially distributed with rate $\mu=0.1$, the batch size is $B=200$, and the total number of requests is $N=200000$.
We run the simulations for 10 different seeds and provide the average throughput and expected latency of the system.
It can be observed that the throughput of the system increases with the number of bins $k$ with the multi-bin batching policy in Figure~\ref{fig:exp_service_time_throughput}.
The average latency of the system depicted in Figure~\ref{fig:exp_service_time_latency} decreases with the number of bins $k$.
It can be seen that with the increasing number of bins, the system first achieves a lower latency but after a certain point, the latency starts to increase.
This is different from the results in the uniform distribution case, where the latency increases with the number of bins.

\section{LLM Experiments}
\label{sec:llm-experiments}

We evaluate multi-bin batching in four complementary settings.
First, we run a controlled real-inference workload using \texttt{llama.cpp}, where output lengths are selected from a balanced uniform-bin distribution and bin assignments use
oracle output lengths.
Second, we evaluate a more realistic Dolly
dataset~\citep{databricks2023dolly} workload, again using oracle output lengths for bin assignment.
Third, we repeat the Dolly workload with bin assignments
obtained from an estimated-length predictor.
Finally, we retain the simulation-based symmetric error study from the original evaluation to isolate the effect of controlled bin-assignment noise.

Throughout this section, throughput denotes generated tokens per second unless otherwise stated.
We use \emph{oracle length} to mean that bin assignment is based on the true generated output length.
We use \emph{estimated length} to mean that bin assignment is based on a length predictor; the inference run itself still generates the true target number of tokens.
This distinction allows us to separate the scheduling benefit available with perfect length information from the benefit attainable with a practical length estimator.

\subsection{Experimental Setup}
\label{sec:llm-experiment-setup}

The real-inference experiments evaluate scheduling policies using \texttt{llama.cpp} as the serving backend.
All real-inference experiments use \texttt{Llama-3.2-3B-Instruct}~\citep{grattafiori2024llama} served through \texttt{llama.cpp}.
For each workload, requests are issued to a local \texttt{llama.cpp} server according to the scheduling policy under evaluation, and throughput is measured as the total number of generated tokens divided by the wall-clock completion time of the workload.
We compare static batching, native continuous batching, and multi-bin static batching.

We compare three scheduling policies. \emph{Static batching} forms fixed batches in request order and does not refill a batch until all requests in that batch complete.
\emph{Continuous batching} uses the backend's native continuous-batching scheduler, which can admit new requests as running requests finish.
This is a substantially stronger and more practically relevant baseline than fixed static batching.
\emph{Multi-bin static batching} assigns requests to length bins, forms fixed batches within each bin, and dispatches ready batches in earliest-request-id order.
Thus, static batching isolates the straggler effect targeted by our method, while the native continuous-batching baseline provides a strong practical reference point for assessing whether length-aware grouping remains useful beyond fixed-batch comparisons.

The controlled oracle-length experiment uses 1024 requests with batch size 32.
Target generation lengths are sampled from a balanced uniform-bin distribution over 500--2000 tokens.
We evaluate $k \in \{2,4,8,16\}$ bins, with a total assigned output length of 1,282,127 tokens.
The Dolly experiments use 1478 test requests from a Dolly-derived labeled workload with batch size 64.
The true target lengths range from 3 to 978 tokens, for a total of 408,352 generated tokens.
All throughput numbers below are reported in generated tokens per second.

All real-inference experiments were run on a single NVIDIA A100 80GB GPU.
The model was served in FP16 precision with a context size of 4096.
Before each measured run, we performed a warm-up run using the same serving configuration.
For the controlled real-inference workload, we use batch size $B=32$.
For the Dolly workload, we use batch size $B=64$.

\subsection{Controlled Oracle-Length Inference}
\label{sec:controlled-oracle-inference}

\begin{figure}[ht]
    \centering
    \includegraphics[width=0.48\textwidth]{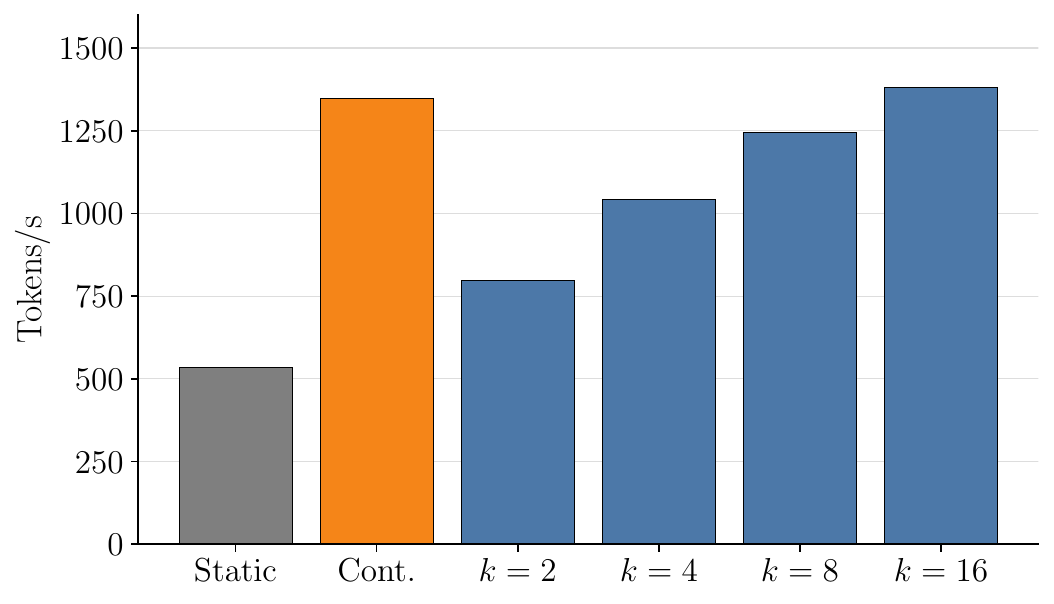}
\caption{Controlled oracle-length \texttt{llama.cpp} throughput.}
    \label{fig:controlled_oracle_llamacpp_throughput}
\end{figure}

Figure~\ref{fig:controlled_oracle_llamacpp_throughput} reports the controlled oracle-length experiment.
Static batching achieves 534.23 tokens/s.
Multi-bin batching improves monotonically with the number of bins, reaching 1380.48 tokens/s at $k=16$.
This corresponds to a 158.4\% improvement over static batching, showing that length-aware grouping can substantially reduce the inefficiency caused by heterogeneous output lengths in fixed batches.

Continuous batching is much stronger than static batching, achieving 1348.49 tokens/s.
In this controlled workload, $k=16$ multi-bin batching is 2.4\% higher than continuous batching, while smaller values of $k$ remain below the continuous baseline.
We view this result as evidence that sufficiently fine oracle-length grouping can approach, and in this case slightly exceed, the throughput of the native continuous-batching baseline under controlled length heterogeneity.

\subsection{Dolly Dataset Inference with Oracle Lengths}
\label{sec:dolly-oracle-inference}

\begin{figure}[ht]
    \centering
    \includegraphics[width=0.48\textwidth]{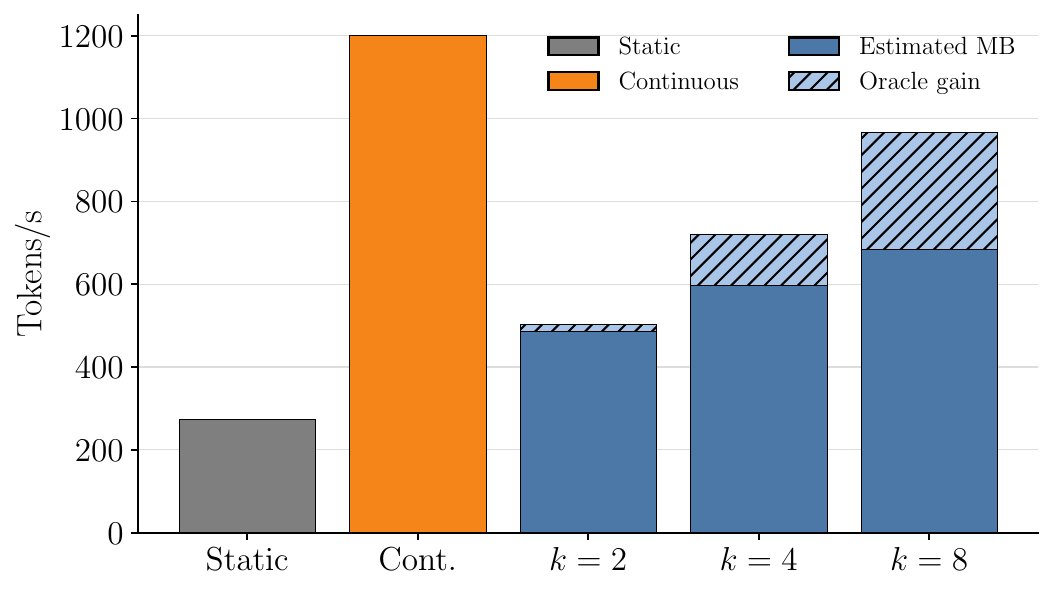}
\caption{Dolly dataset~\citep{databricks2023dolly} llama.cpp throughput with estimated and oracle output lengths. For multi-bin bars, the solid portion shows estimated-length batching, and the hatched segment shows the additional throughput obtained with oracle lengths.}
    \label{fig:dolly_oracle_estimated_llamacpp_throughput}
\end{figure}

Figure~\ref{fig:dolly_oracle_estimated_llamacpp_throughput} reports the Dolly real-inference experiment with oracle bin assignments.
Because bins are assigned using the true output lengths, this experiment isolates the scheduling benefit from any prediction error.
Accordingly, these oracle results should be interpreted as an upper bound on the scheduling gains available from accurate length information, rather than as directly achievable without a sufficiently accurate estimator.
Static batching achieves 273.53 tokens/s, whereas continuous batching achieves 1200.07 tokens/s.
Multi-bin batching improves over static batching for all tested values of $k$: $k=2$ achieves 502.65 tokens/s, $k=4$ achieves 720.17 tokens/s, and $k=8$ achieves 965.97 tokens/s.

Relative to static batching, the $k=8$ oracle result gives a 253.1\% throughput improvement.
However, unlike the controlled oracle-length workload, the Dolly oracle result remains below the native continuous-batching baseline: $k=8$ oracle multi-bin batching is 19.5\% below continuous batching.
These results emphasize two points.
First, accurate length information can recover a large amount of the static-batching inefficiency.
Second, the remaining gap to continuous batching suggests that more fine-grained binning, adaptive bin boundaries, or integration with continuous-batching mechanisms may be needed to fully exploit length information.

\subsection{Dolly Dataset Inference with Estimated Lengths}
\label{sec:dolly-estimated-inference}

\begin{table}[ht]
\centering
\begin{tabular}{ccc}
\toprule
Bins & Exact acc. & Adjacent acc. \\
\midrule
$k=2$ & 0.943 & 1.000 \\
$k=4$ & 0.791 & 0.998 \\
$k=8$ & 0.567 & 0.949 \\
\bottomrule
\end{tabular}
\caption{Estimator bin-assignment quality on the Dolly dataset.}
\label{tab:dolly_estimator_accuracy}
\end{table}

We next evaluate the same Dolly workload when bin assignments are obtained from estimated rather than oracle output lengths.
The estimator is a lightweight length regressor built on prompt-prefill hidden-state features from Llama-3.2-3B-Instruct, motivated by recent work showing that the served LLM's internal representations can provide useful signal for output-length prediction~\citep{xie2026predicting}.
Each prompt is processed with the same Llama-3 style chat template used for label generation, but no answer tokens are generated
for bin assignment.
This design aligns the predictor with the model used in the serving experiments, while keeping bin assignment limited to prompt-prefill computation.

\begin{figure*}[ht]
    \centering
    \includegraphics[width=0.95\textwidth]{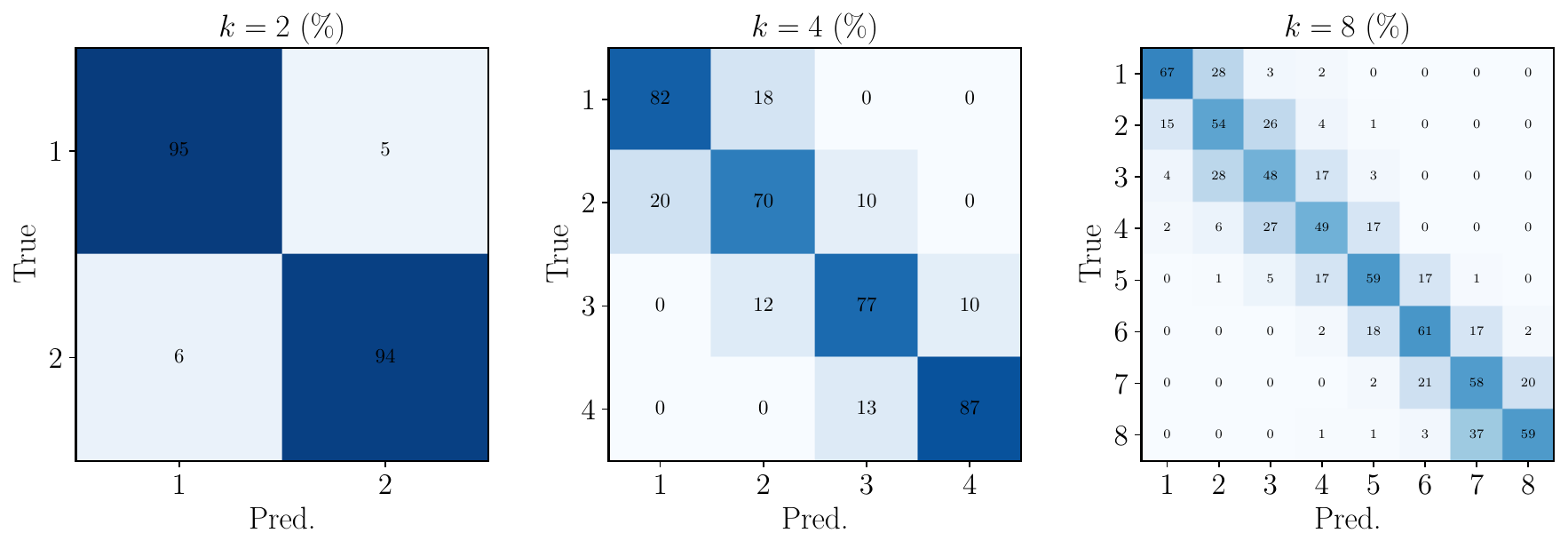}
\caption{Row-normalized confusion matrices for the estimated Dolly dataset bin assignments.}
    \label{fig:dolly_estimator_confusion_matrices}
\end{figure*}

Table~\ref{tab:dolly_estimator_accuracy} summarizes the resulting bin-assignment quality.
The realized exact bin accuracies are 94.28\%, 79.09\%, and 56.69\% for $k=2$, $k=4$, and $k=8$, respectively.
The corresponding adjacent-bin accuracies are 100.0\%, 99.80\%, and 94.89\%.
Thus, although exact bin accuracy decreases as the number of bins increases, most errors remain localized near the correct bin, especially for $k=8$, where fewer than 5.2\% of requests are assigned more than one bin away from their oracle bin.
Figure~\ref{fig:dolly_estimator_confusion_matrices} shows the corresponding row-normalized confusion matrices; for readability, the $k=8$ panel is annotated using whole-number percentages.

As shown in Figure~\ref{fig:dolly_oracle_estimated_llamacpp_throughput}, estimated-length binning still improves substantially over static batching.
Throughput increases from 273.53 tokens/s under static batching to 485.70, 597.91, and 684.02 tokens/s for $k=2$, $k=4$, and $k=8$, respectively.
The best estimated-length result, obtained at $k=8$, is 150.1\% higher than static batching.
However, it remains 43.0\% below the continuous-batching baseline.

These results indicate that practical length estimation preserves a large fraction of the gain over static batching, while leaving substantial room for further improvement.
The gap between the estimated and oracle bars in Figure~\ref{fig:dolly_oracle_estimated_llamacpp_throughput} quantifies the remaining opportunity from more accurate length information: at $k=8$, oracle assignments increase throughput from 684.02 to 965.97 tokens/s.
Although the Dolly oracle result remains below the continuous-batching baseline in this experiment, it closes much of the gap and suggests that finer binning, improved lightweight length estimators, and bin boundaries adapted to workload-specific prediction errors can further strengthen multi-bin batching.
These results support our view that length-aware grouping is a useful scheduling principle that can complement modern continuous-batching systems.

\subsection{Effect of Symmetrical Prediction Errors}
\label{sec:symmetric-error-results}

Finally, we retain the controlled robustness study with symmetric binning errors.
This experiment is a simulation rather than a measured \texttt{llama.cpp} inference run.
Its purpose is to isolate how multi-bin batching degrades as bin assignments become noisier under a simple and controlled error model.

For a true interior bin $i$, the predicted bin is $i-1$ with probability $p_e$, $i+1$ with probability $p_e$, and $i$ with probability $1-2p_e$.
For edge bins, the prediction moves to the only neighboring bin with probability $p_e$ and remains correct with probability $1-p_e$.
This error model captures adjacent-bin mistakes, which are also the dominant type of error observed in the estimated Dolly experiments.
It should not be interpreted as a full model of real predictor errors, which may be asymmetric, workload-dependent, or span multiple bins.

\begin{figure*}[ht]
    \centering
    \begin{subfigure}[ht]{0.32\textwidth}
        \centering
        \includegraphics[width=\textwidth]{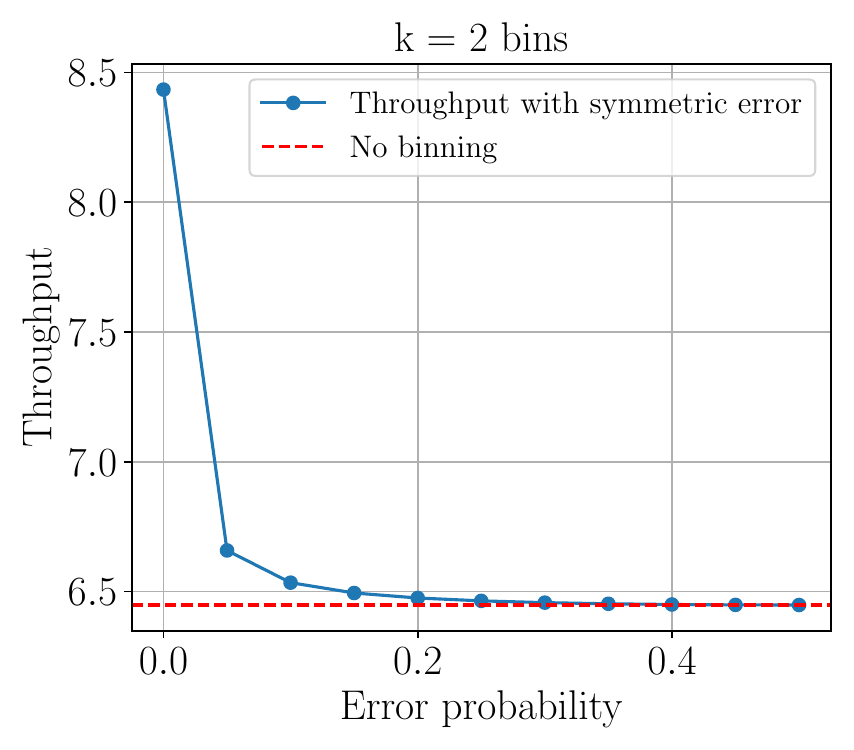}
        \caption{2 bins}
        \label{fig:throughput_vs_error_prob_k_2}
    \end{subfigure}
    \begin{subfigure}[ht]{0.32\textwidth}
        \centering
        \includegraphics[width=\textwidth]{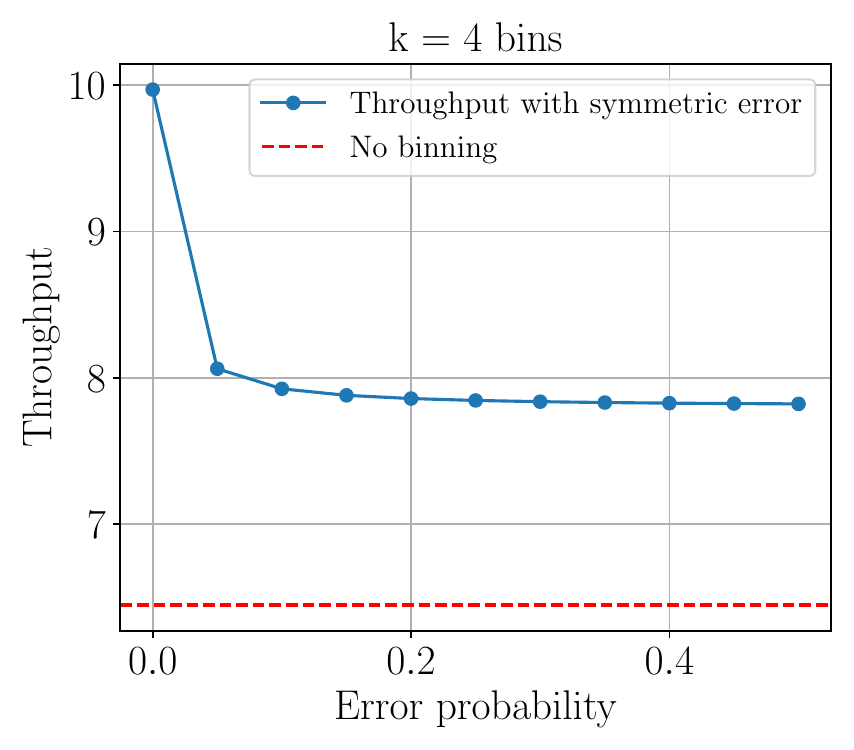}
        \caption{4 bins}
        \label{fig:throughput_vs_error_prob_k_4}
    \end{subfigure}
    \begin{subfigure}[ht]{0.32\textwidth}
        \centering
        \includegraphics[width=\textwidth]{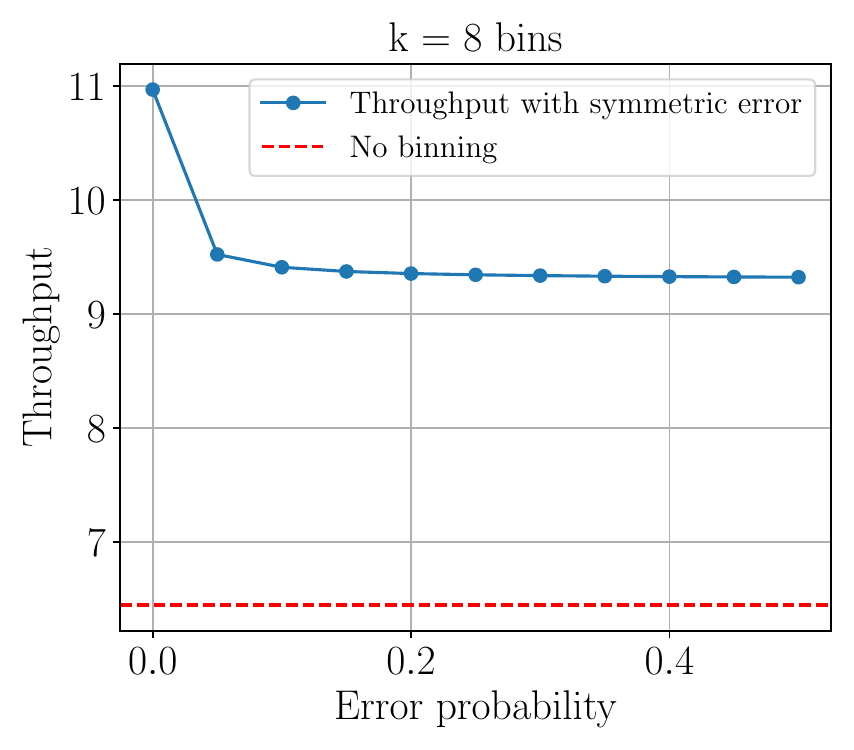}
        \caption{8 bins}
        \label{fig:throughput_vs_error_prob_k_8}
    \end{subfigure}
\caption{Throughput as a function of the symmetric adjacent-bin error probability $p_e$ for different values of $k$.}
    \label{fig:throughput_vs_error_prob}
\end{figure*}

Figure~\ref{fig:throughput_vs_error_prob} reports throughput as $p_e$ varies for $k=2$, $k=4$, and $k=8$.
The simulation uses batch size $B=128$, service times between $l_{\min}=1$ and $l_{\max}=20$, and 128000 submitted requests.
Results are averaged over 10 simulation runs.
As expected, throughput decreases as the error probability increases.
Nevertheless, the multi-bin policy remains robust to moderate adjacent-bin errors.
This simulation complements the estimated Dolly experiment by showing, in a controlled setting, how throughput changes as the binning signal becomes progressively noisier.

\section{Conclusion}
This work presented multi-bin batching, a queueing-based control policy that groups requests with similar output lengths to improve batching efficiency.
Our theoretical results show that length-aware binning improves the effective batched service rate and approaches the idealized static-batching capacity as the number of bins increases.
The analysis isolates the effect of output-length heterogeneity and should be interpreted as complementary to, rather than a complete model of, modern continuous-batching systems.
Our empirical studies support this interpretation. The revised real-inference experiments show significant gains over static batching, and they also compare directly with the backend's native continuous-batching behavior.
On the Dolly workload, both oracle and estimated-length multi-bin batching substantially improve over static batching. Although they remain below the backend's native continuous-batching baseline in this experiment, the oracle $k=8$ result narrows the gap considerably, and the estimated-length result demonstrates that practical length prediction still preserves significant throughput gains.
These results suggest that accurate length information can provide a useful scheduling signal, while also showing that predictor quality is central to the practical value of the method.
Rather than replacing modern serving systems, multi-bin batching is best viewed as a length-aware grouping principle that may complement them through queue classes, admission-control decisions, or replica-routing policies.
Future work could focus on improving lightweight length predictors, incorporating uncertainty-aware bin assignment, developing adaptive binning strategies, and studying the full latency--throughput Pareto tradeoff in end-to-end serving systems.

\subsubsection*{Acknowledgments}

Ozgur Guldogan and Ramtin Pedarsani were supported by the National Science Foundation (NSF) under Awards ECCS-2419982 and CCF-2342253.
Kangwook Lee was supported in part by the National Science Foundation (NSF) CAREER Award CCF-2339978, an Amazon Research Award, and a grant from FuriosaAI.
Jackson Kunde was supported by the University of Wisconsin--Madison Hilldale Undergraduate Research Fellowship (2024 cycle).

\bibliographystyle{plainnat}
\bibliography{llm_scheduling.bib}

\appendix
\section{Appendix}
\subsection{Proof of Lemma~\ref{lemma:latency}}\label{sec:proof_latency}
In this section, we provide the proof of Lemma~\ref{lemma:latency}.
\begin{proof}
    Under the assumption of infinite servers, the latency consists of the time spent waiting to complete the current batch and the service time.
    Therefore, the expected latency of a request is given by
    \begin{equation}
        \mathbb{E}[t_{\text{latency}}] = \mathbb{E}[t_{\text{batch}}] + \mathbb{E}[t_{\text{service}}].
    \end{equation}
    The expected time spent waiting to complete the current batch is given by
    \begin{equation}
        \mathbb{E}[t_{\text{batch}}] = \sum_{i=1}^{k} \mathbb{P}(\text{bin}=i)\mathbb{E}[t_{\text{batch}}| \text{bin}=i].
    \end{equation}
    Since the bins are equally likely, the arrival rate for each bin is $\lambda/k$.
    Then, for each request in the batch, the expected time spent waiting to complete the current batch is given by
    \begin{equation}
        \mathbb{E}[t_{\text{batch}}| \text{bin}=i] = \frac{1}{B}\sum_{j=1}^{B} \frac{(B-j)k}{\lambda} = \frac{B-1}{2\lambda}k.
    \end{equation}
    Then, the expected time spent waiting to complete the current batch is given by
    \begin{equation}
        \mathbb{E}[t_{\text{batch}}] = \frac{B-1}{2\lambda}k.
    \end{equation}
    The expected service time of a request is given by Theorem~\ref{thm:throughput_k_bin}.
    Therefore, the expected latency of a request can be derived as in the statement of the lemma.
\end{proof}

\subsection{Exponentially Distributed Service Time}\label{sec:proof_exp_service_time}

In this section, we provide the proof of Lemma~\ref{lem:exp_service_time}.
\begin{proof}
    The upper bound on the expected service time of a batch of $B$ requests with $k$ bins is derived based on the following observation.
    \begin{equation}
        \E[t_{\text{service, k}}| \text{bin}=i] \leq l_{i} \quad \forall i\in[k-1],
    \end{equation}
    and the exact expected service time of a batch of $B$ requests with $k$ bins is given by $l_{k-1}+\frac{H_B}{\mu}$.
    This is well-known in the literature, it is the maximum of shifted exponential random variables.
    Then, the expected service time of a batch of $B$ requests with $k$ bins is upper bounded by the sum of the expected service time of each bin.
    The upper bound could be written as follows,
    \begin{equation}
        \E[t_{\text{service, k}}] \leq \sum_{i=1}^{k-1} \Pr(\text{bin}=i)l_{i} + \Pr(\text{bin}=k)\left(l_{k-1} + \frac{H_B}{\mu}\right).
    \end{equation}
    The probability of each bin is given by $\Pr(\text{bin}=i) = \exp(-\mu l_{i-1}) - \exp(-\mu l_{i})$.
    Then, we apply the following change of variables before minimizing the upper bound.
    Let $q_{i} = \exp(-\mu l_{i})$ and $q_0=1$, then the upper bound can be written as,
    \begin{equation}
        \E[t_{\text{service, k}}] \leq \sum_{i=1}^{k-1} (q_{i-1} - q_{i})\frac{\log(1/q_{i})}{\mu} + q_{k-1}\left(\frac{\log(1/q_{k-1})}{\mu} + \frac{H_B}{\mu}\right) = f(\boldsymbol{q}).
    \end{equation}
    It can be seen the upper bound function can be decomposed as a function of $\boldsymbol{q}=[q_{1}, q_{2}, \ldots, q_{k-1}]$ and a multiplicative factor of $1/\mu$.
    Therefore, we will assume that $\mu=1$ for simplicity.
    Then, we can write the upper bound function as,
    \begin{align}
        f(\boldsymbol{q}) &= \sum_{i=1}^{k-1} (q_{i-1} - q_{i})\log(1/q_{i}) + q_{k-1}\left(\log(1/q_{k-1}) + H_B\right) \\
        &= \sum_{i=1}^{k-1} (q_i-q_{i-1})\log(q_{i}) + q_{k-1}\left(H_B - \log(q_{k-1})\right).
    \end{align}
    We can compute the partial derivative of $f(\boldsymbol{q})$ with respect to $q_{i}, i\in[k-2]$ as follows,
    \begin{equation}
        \frac{\partial f(\boldsymbol{q})}{\partial q_{i}} = \log(q_{i}) + \frac{q_{i}-q_{i-1}}{q_{i}} - \log(q_{i+1}) \quad \forall i\in[k-2].
    \end{equation}
    Then, the partial derivative with respect to $q_{k-1}$ is given by,
    \begin{equation}
        \frac{\partial f(\boldsymbol{q})}{\partial q_{k-1}} = H_B - \frac{q_{k-2}}{q_{k-1}}.
    \end{equation}
    Then, the second-order partial derivative of $f(\boldsymbol{q})$ with respect to $q_{i}q_{j}, i,j\in[k-1]$ is given by,
    \begin{equation}
        \frac{\partial^{2} f(\boldsymbol{q})}{\partial q_{i}\partial q_{j}} = \begin{cases}
            \frac{1}{q_{i}} + \frac{q_{i-1}}{q_{i}^{2}} & \text{if } i=j \text{ and } i\in[k-2],\\
            \frac{q_{i-1}}{q_{i}^{2}} & \text{if } i=j=k-1,\\
            -\frac{1}{q_{\max(i,j)}} & \text{if } |i-j|=1,\\
            0 & \text{otherwise}.
        \end{cases}
    \end{equation}
    Then, the Hessian matrix of $f(\boldsymbol{q})$ is a tridiagonal matrix in the form of,
    \begin{equation}
        \nabla^{2} f(\boldsymbol{q}) = \begin{bmatrix}
            \frac{q_{1}+q_{0}}{q_{1}^{2}} & -\frac{1}{q_{2}}  & \cdots & 0 & 0\\
            -\frac{1}{q_{2}} & \frac{q_{2}+q_{1}}{q_{2}^{2}} & \cdots & 0 & 0\\
            0 & -\frac{1}{q_{3}} & \cdots & 0 & 0\\
            \vdots & \vdots & \ddots & \vdots & \vdots\\
            0 & 0 & \cdots & \frac{q_{k-1}+q_{k-2}}{q_{k-1}^{2}} & -\frac{1}{q_{k-1}} \\
            0 & 0 & \cdots & -\frac{1}{q_{k-1}} & \frac{q_{k-2}}{q_{k-1}^{2}}
        \end{bmatrix}.
    \end{equation}
    The determinant of the matrix can be found using the following recursive formula:
    \begin{equation}
        f_n = A_{n,n}f_{n-1} - A_{n,n-1}A_{n-1,n}f_{n-2} \quad \forall n \in [2,k-1]
    \end{equation}
    where $f_1 = A_{1,1}$ and $f_0 = 1$.
    For all $n \in [2,k-2]$, it can be seen that:
    \begin{equation}
        f_n = \frac{q_{n-1} + q_n}{q_n^2}f_{n-1} - \frac{1}{q_{n}^2}f_{n-2}
    \end{equation}
    Our claim is that:
    \begin{equation}
        f_n = \frac{1}{q_1 q_2 \ldots q_{n-1}q_n^2} + \frac{1}{q_n}f_{n-1} \quad \forall n \in [2,k-2]
    \end{equation}
    It holds for $n=1$.
    Then, we can prove it by induction.
    Assume that it holds for $n-1$.
    Then, we can write the following:
    \begin{align}
        f_n &= \frac{q_{n-1} + q_n}{q_n^2}f_{n-1} - \frac{1}{q_n^2 }f_{n-2} \\
        &= \frac{q_{n-1} + q_n}{q_n^2}\left(\frac{1}{q_1 q_2 \ldots q_{n-2}q_{n-1}^2} + \frac{1}{q_{n-1}}f_{n-2}\right) - \frac{1}{q_n^2}f_{n-2} \\
        &= \frac{1}{q_1 q_2 \ldots q_{n-1}q_n^2} + \frac{1}{q_n}f_{n-1}
    \end{align}
    Hence, it is proven by induction.

    Then, we can compute the determinant of the Hessian as:
    \begin{equation}
        \det(\nabla^{2} f) = f_{k-1} = \frac{q_{k-2}}{q_{k-1}^2}f_{k-2} - \frac{1}{q_{k-1}^2}f_{k-3}
    \end{equation}
    We can replace the $f_{k-2}$ with the formula:
    \begin{equation}
        f_{k-2} = \frac{1}{q_1 q_2 \ldots q_{k-3}q_{k-2}^2} + \frac{1}{q_{k-2}}f_{k-3}
    \end{equation}
    Then, we can compute the determinant of the matrix as:
    \begin{align}
        \det(\nabla^{2} f) &= \frac{q_{k-2}}{q_{k-1}^2}\left(\frac{1}{q_1 q_2 \ldots q_{k-3}q_{k-2}^2} + \frac{1}{q_{k-2}}f_{k-3}\right) - \frac{1}{q_{k-1}^2}f_{k-3}\\
        &= \frac{1}{q_1q_2\ldots q_{k-2}q_{k-1}^2}
    \end{align}
    Therefore, the determinant of the Hessian matrix is as follows:
    \begin{equation}
        \det(\nabla^2 f) = \frac{1}{q_1q_2\ldots q_{k-2}q_{k-1}^2}>0
    \end{equation}
    Since all $q_i$ are positive, the recursion above shows that all leading principal minors of the Hessian are positive.
    Therefore, by Sylvester's criterion, the Hessian is positive definite, and the upper bound for the total service time is a convex function of the transformed decision variables $\boldsymbol{q}$.
    Then, the optimal decision boundaries can be found by setting the partial derivative of the upper bound function to zero.
    We can start from the partial derivative with respect to $q_{k-1}$ as follows:
    \begin{equation}
        \frac{\partial f}{\partial q_{k-1}} = H_B - \frac{q_{k-2}}{q_{k-1}} = 0 \implies q_{k-2} = q_{k-1}H_B
    \end{equation}
    Then, we can compute the partial derivative with respect to $q_{i}, i\in[k-2]$ as follows:
    \begin{align}
        &\frac{\partial f}{\partial q_{i}} = \log(q_{i}) + \frac{q_{i}-q_{i-1}}{q_{i}} - \log(q_{i+1}) = 0 \implies \frac{q_{i-1}}{q_{i}} = 1+ \log\left(\frac{q_{i}}{q_{i+1}}\right)
    \end{align}
    Utilizing the above equation and $q_0=1$, we can derive,
    \begin{align}
        \frac{1}{q_1} &= 1 + \log\left(\frac{q_1}{q_2}\right)\\
        &= 1 + \log\left(1+ \log\left(\frac{q_2}{q_3}\right)\right)\\
        &= 1 + \log\left(1+ \log\left(1+ \ldots + \log\left(\frac{q_{k-2}}{q_{k-1}}\right)\right)\right)\\
        &= 1 + \log\left(1+ \log\left(1+ \ldots + \log(H_B)\right)\right) = L_{k-1} \implies  q_1 = \frac{1}{L_{k-1}}
    \end{align}
    where $L_{k-1}$ is defined recursively as:
    \begin{equation}
    L_m = \begin{cases}
    H_B & \text{if } m = 1 \\
    1+ \log(L_{m-1}) & \text{if } m > 1
    \end{cases}
    \end{equation}
    Similarly, we can derive $q_{i}$ for $i\in[k-1]$ as follows:
    \begin{equation}
    q_{i} = \frac{1}{\prod_{j=1}^{i}L_{k-j}} \quad \forall i\in[k-1]
    \end{equation}
    Then, the optimal decision boundaries are given by,
    \begin{equation}
    l_{i} = -\frac{1}{\mu}\log(q_{i}) = \frac{1}{\mu}\sum_{j=1}^{i}\log(L_{k-j}) \quad \forall i\in[k-1].
    \end{equation}
    This completes the proof.
\end{proof}

\end{document}